\documentclass[letterpaper, 10 pt, conference]{ieeeconf}  %

\IEEEoverridecommandlockouts                              %

\usepackage{graphics} %
\usepackage{epsfig} %
\usepackage{mathptmx} %
\usepackage{times} %
\usepackage{amsmath} %
\usepackage{amssymb}  %
\usepackage{multirow}
\usepackage{multicol}
\usepackage{algorithm}
\usepackage{algpseudocode}
\usepackage{mathrsfs}
\usepackage{xcolor}
\usepackage{dblfloatfix}
\usepackage{hyperref}

\newcommand{\sysName}{MatchMaker}

\title{\LARGE \bf

MatchMaker: Automated Asset Generation for Robotic Assembly 
}

\author{Yian Wang, Bingjie Tang, Chuang Gan, Dieter Fox, Kaichun Mo, Yashraj Narang, Iretiayo Akinola}%

\begin{document}

\setcounter{figure}{1}
\makeatletter
\let\@oldmaketitle\@maketitle%
\renewcommand{\@maketitle}{
   \@oldmaketitle%
   \begin{center}
    \centering      
    \noindent\includegraphics[width=0.95\linewidth]{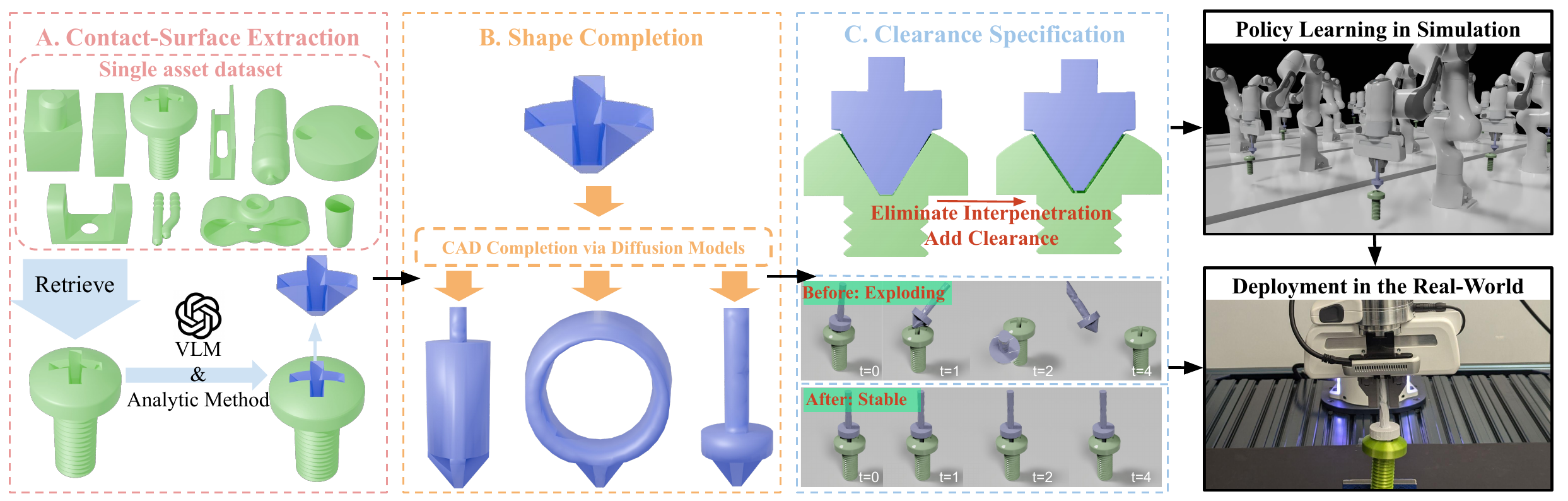}\qquad
  \end{center}
  \footnotesize{\textbf{Fig.~\thefigure:\label{fig:teaser}}~ \textbf{{\sysName}} automatically generates assembly pairs in three stages:  
1) Contact-Surface Detection: Identifies the assembly axis and contact faces of an initially sampled asset for the assembly pair.  
2) Shape Completion: Utilizes diffusion models to complete the rest of the asset, generating the second asset of the pair.  
3) Clearance Specification: Erodes the contacting surfaces to prevent interpenetration and achieve the desired clearance.
We further demonstrate that our generated asset pairs are simulatable by training assembly policies within NVIDIA Isaac Simulator and realizable by deploying the learned policies with 3D-printed assets in the real world.
  }
\vspace{-2mm}
}
\makeatother 

\maketitle
\thispagestyle{empty}
\pagestyle{empty}

\begin{abstract}
Robotic assembly remains a significant challenge due to complexities in visual perception, functional grasping, contact-rich manipulation, and performing high-precision tasks. Simulation-based learning and sim-to-real transfer have led to recent success in solving assembly tasks in the presence of object pose variation, perception noise, and control error; however, the development of a generalist (i.e., multi-task) agent for a broad range of assembly tasks has been limited by the need to manually curate assembly assets, which greatly constrains the number and diversity of assembly problems that can be used for policy learning. Inspired by recent success of using generative AI to scale up robot learning, we propose {\sysName}, a pipeline to automatically generate diverse, simulation-compatible assembly asset pairs to facilitate learning assembly skills. Specifically, {\sysName} can 1) take a simulation-incompatible, interpenetrating asset pair as input, and automatically convert it into a simulation-compatible, interpenetration-free pair, 2) take an arbitrary single asset as input, and generate a geometrically-mating asset to create an asset pair, 3) automatically erode contact surfaces from (1) or (2) according to a user-specified clearance parameter to generate realistic parts.
We demonstrate that data generated by {\sysName} outperforms previous work in terms of diversity and effectiveness for downstream assembly skill learning. Project page: \url{https://wangyian-me.github.io/MatchMaker/}.

\end{abstract}

\section{INTRODUCTION}

Contact-rich assembly tasks are prevalent in real-world applications, including household tasks like phone charging, USB connections, furniture assembly, and various industrial tasks \cite{whitney2004mechanical} such as assembling nuts and bolts, inserting bearings, and fastening components in automotive production.
Although these tasks have relatively concrete goals, typically involving moving two assets into a fixed relative pose, they present significant challenges for autonomous robots. Robots must accurately perceive, grasp, and insert objects with precision to complete these tasks, often under varying environmental conditions and uncertainties in object positioning.
To address these challenges, recent works have developed fast simulations for contact-rich scenarios \cite{narang2022factory}, sim-to-real methods for transferring contact-rich assembly policies trained in simulation to the real world \cite{schoettler2020meta, davchev2022residual, zhang2022learning, kozlovsky2022reinforcement, allshire2022transferring, tang2023industreal}, and assembly policy-learning algorithms capable of handling a single asset pair \cite{tang2024automate}. However, training a general policy that can reliably assemble a wide range of assets—whether seen or unseen—with diverse geometries remains an unsolved problem. The most capable generalist to date achieves an 80 percent success rate on 20 seen assemblies \cite{tang2024automate}, but these represent only a small fraction of the diversity found in real-world assemblies.

As highlighted in several previous works \cite{bharadhwaj2024roboagent, ehsani2023imitating, ebert2021bridge, padalkar2023open, brohan2022rt, brohan2023rt}, scaling up data significantly enhances the generalization capabilities of robot agents, leading to promising results in acquiring diverse skills across multiple scenes, tasks, and learning algorithms.
Similarly, related studies \cite{mandlekar2023mimicgen, ankile2024juicer, wang2023gensim, wang2023robogen} have demonstrated the value of automatically collecting large-scale demonstration data in simulated environments. Also, recent works \cite{heo2023furniturebench, ankile2024juicer} have introduced methods for automatically collecting demonstrations for two-part or multi-part assembly tasks. Combining these approaches with sim-to-real methods \cite{tang2023industreal} opens the possibility of developing agents capable of solving a variety of assembly tasks with diverse geometries. However, the limited availability of assembly asset pairs (e.g., different plug-and-socket combinations) remains a significant obstacle.

Previous efforts \cite{tian2022assemble, willis2022joinable} have collected larger datasets with several thousand assets for contact-rich policy learning. However, these assets are not entirely penetration-free or have insufficient clearance between paired components (i.e., in the assembled state, elements of the meshes of two mating parts intersect each other). This makes them incompatible with many high-accuracy simulators which often attempt to solve strict non-penetration constraints. In addition, they are not feasible in the real world, as the parts cannot be physically assembled after manufacturing (e.g. 3D printing). 
Previous work, AutoMate \cite{tang2024automate}, introduced the largest dataset that includes positive clearances and is compatible with most simulators. However, the assets were manually adjusted based on \cite{tian2022assemble}, requiring approximately 40 hours of human effort to process 100 assets. All previous datasets have been collected through manual effort, making it nearly impossible to scale them up efficiently. These limitations highlight the need for a scalable approach to generate large assembly datasets with minimal human effort.

To address these challenges, we propose {\sysName}, a pipeline that automatically generates diverse, simulation-compatible paired assets for single-axis assembly tasks using a diffusion-based shape completion approach with controllable clearances between assets. Given that generating a single asset is significantly easier than generating paired assets (as evidenced by the large dataset with 1 million assets \cite{koch2019abc} and extensive research in this area \cite{xu2024brepgen, jayaraman2022solidgen, wu2021deepcad}), 
we aim to leverage the single-asset generation process to achieve paired-asset generation. Unlike generating two individual assets, paired-asset generation requires that the two assets can seamlessly assemble, i.e., share several contact surfaces. 
To ensure proper alignment, we enforce geometric constraints on contact regions, using identified contact surfaces as seeds for the generation process. A shape completion step then generates the full paired assets while preserving these constraints.

Specifically, our pipeline consists of three steps, as illustrated in \textbf{Fig. 1}:
1) \textbf{Contact-Surface Extraction:} we begin with a single asset sampled from an existing dataset or generated individually and use a large vision-language model \cite{islam2024gpt} to identify the assembly axis along which an insertion operation can occur. Then we identify the contacting surfaces that will interface with the paired/second asset using an analytical method. 
2) \textbf{Shape Completion:} After that, we execute a shape completion step, where a 3D generative model \cite{xu2024brepgen} fills out the regions outside the contacting surfaces to generate the full paired asset. 
3) \textbf{Clearance Specification:} 
Finally, a post-processing step erodes the contacting region of the paired asset to achieve a parameterized level of clearance, eliminating any possible interpenetration and ensuring suitability for assembly tasks. This control over clearance can for real-world variability in part clearances and tolerances, which directly affect the forces, precision, and difficulty involved in assembly.
Once the paired assets are generated, they can be loaded into a simulator for skill learning.

We summarize our main contributions as follows:
\begin{itemize}
    \item We introduce {\sysName}, a generative pipeline that creates diverse and realistic paired assets for assembly skill learning.
    \item We formulate the paired-asset generation problem as a shape completion task, leveraging large vision-language models to identify assembly axes, an analytical method to detect shared surfaces, and a set of 3D diffusion models to complete the generation of paired assets.
    \item We develop a clearance-specification tool that adjusts paired assets to control inter-asset clearance and ensure penetration-free interactions.
    \item We validate our method through experiments in robot policy learning using our generated assets in both simulation and the real-world, demonstrating the effectiveness of the proposed approach.
\end{itemize}
To the best of our knowledge, this work is the first to generate diverse and simulatable paired assets for robotic assembly tasks. 
Our goal is to create opportunities for the research community to develop reliable policies that can generalize across a wide range of assembly tasks. We will release the assembly asset-generation code upon publication.

\section{RELATED WORK}

\subsection{Policy Learning for Robotic Assembly}
There have been a number of recent advancements in simulating contact-rich interactions \cite{narang2022factory, makoviychuk2021isaac}, policy learning within simulation \cite{makoviychuk2021isaac, hansen2022temporal} or the real world \cite{zhao2022offline, spector2021insertionnet, spector2022insertionnet}, and transferring policies from simulation to the real world \cite{schoettler2020meta, davchev2022residual, zhang2022learning, kozlovsky2022reinforcement, allshire2022transferring, tang2023industreal}. %
For assembly, such efforts typically involve learning specialized policies for mating specific assets, with limited consideration for generalization across different shapes and sizes. More recently, some works have begun exploring the generalization of contact-rich skills across varying scales and sizes \cite{huang2021generalization}. A recent study \cite{tang2024automate} focused on learning a generalist (i.e., multi-task) policy capable of handling dozens of different assembly assets. 
However, since the dataset in \cite{tang2024automate} contained only 100 assets, the ability to learn broadly-generalizable skills remains limited.

\subsection{Datasets for Robotic Assembly}
Numerous 3D asset datasets are commonly used for robotic manipulation research \cite{chang2015shapenet, deitke2023objaverse, deitke2024objaverse}, but most are designed for grasping and pick-and-place applications with limited contact between parts. Fewer datasets address the growing need for planning and skill learning in robotic assembly \cite{jones2021automate, koch2019abc, willis2021fusion, ebinger2018general, zhang2020c}. Although some of these datasets are quite large, they consist of single assets with rather than paired assets for assembly tasks \cite{koch2019abc}. There are also datasets that contain assemblies \cite{tian2022assemble, willis2022joinable, jones2021automate}, but they exhibit issues that make them difficult to disassemble or assemble. \cite{tian2022assemble} post-processed a subset of these datasets to create unique watertight meshes, resulting in the largest dataset for assembly planning research with 12970 assets. However, more recent work \cite{tang2024automate} noted that this dataset was incompatible with some state-of-the-art (SOTA) simulators due to intersection between assets in the assembled state, requiring manual post-processing to remove such artifacts.

\subsection{3D Shape Generation}
\label{section:2b}
Generative models have proven highly effective in content generation, including image generation \cite{chang2022maskgit}, text generation \cite{achiam2023gpt}, and audio generation \cite{kreuk2022audiogen}. This progress is now extending to 3D asset generation in various forms and formats. 
Some works directly generate meshes as the network output \cite{chen2020bsp, siddiqui2024meshgpt, chen2024meshanything}, whereas other works leverage diffusion models to generate 3D shapes \cite{nichol2022point, jun2023shap, qiu2024richdreamer, xu2024instantmesh}. 
In our work, since we are mostly dealing with industrial assets with smooth and relatively simple geometries, we focus on generating CAD models \cite{wu2021deepcad, nash2020polygen, jayaraman2022solidgen, xu2024brepgen}, specifically B-rep CAD representations. 
B-rep is particularly interesting as it is a de-facto standard in CAD modeling, offering a mathematically-precise representation of 3D objects with valuable properties like smoothness and the ability to retain features and topology during zooming or scaling \cite{ansaldi1985geometric}. 
Furthermore, some of the largest 3D asset datasets are available in the B-rep format \cite{koch2019abc, willis2021fusion, jayaraman2021uv, jones2021automate, lambourne2021brepnet}. To the best of our knowledge, this work is the first to use generative models for assembly asset generation, incorporating geometric constraints to ensure that parts can be correctly mated.

\section{Method}

In general, {\sysName} accepts three types of input: 1) no input, 2) a single asset, or 3) an assembly asset pair, and outputs a simulation-compatible assembly asset pair with configurable clearance.
An assembly asset pair consists of two 3D geometries (typically a \textit{plug}, the part to be inserted, and a \textit{receptacle}, the mating part) that must be mated in a specific goal configuration that satisfies certain geometric constraints to enable successful assembly.

We formulate this task as a shape completion problem, where given one asset $A$ of the assembly, the other part $B$ is generated to satisfy the required assembly constraints.
With one input or generated asset (if nothing is input), {\sysName} generates a variety of matching assets that can be assembled with the first part. {\sysName} operates in three main stages, as shown in  \textbf{Fig. 1}:
1) Contact Surface Extraction, which combines high-level reasoning from a vision-language model (e.g., GPT-4o \cite{islam2024gpt}) with an analytical method to identify contact faces on the initial part; 2) Shape Completion, where the contact faces constrain a 3D generative model \cite{xu2024brepgen} to produce the second part; and 3) Clearance Specification, which refines the asset pair by eroding contact surfaces to ensure realism and compatibility with simulators. 
If a pair of assembly assets is provided, only the third stage is applied.

\subsection{Contact-Surface Extraction}
\label{section:A}

\begin{figure}[htb]
\vspace{-2mm}
\begin{center}
\noindent\includegraphics[width=1.02\linewidth]{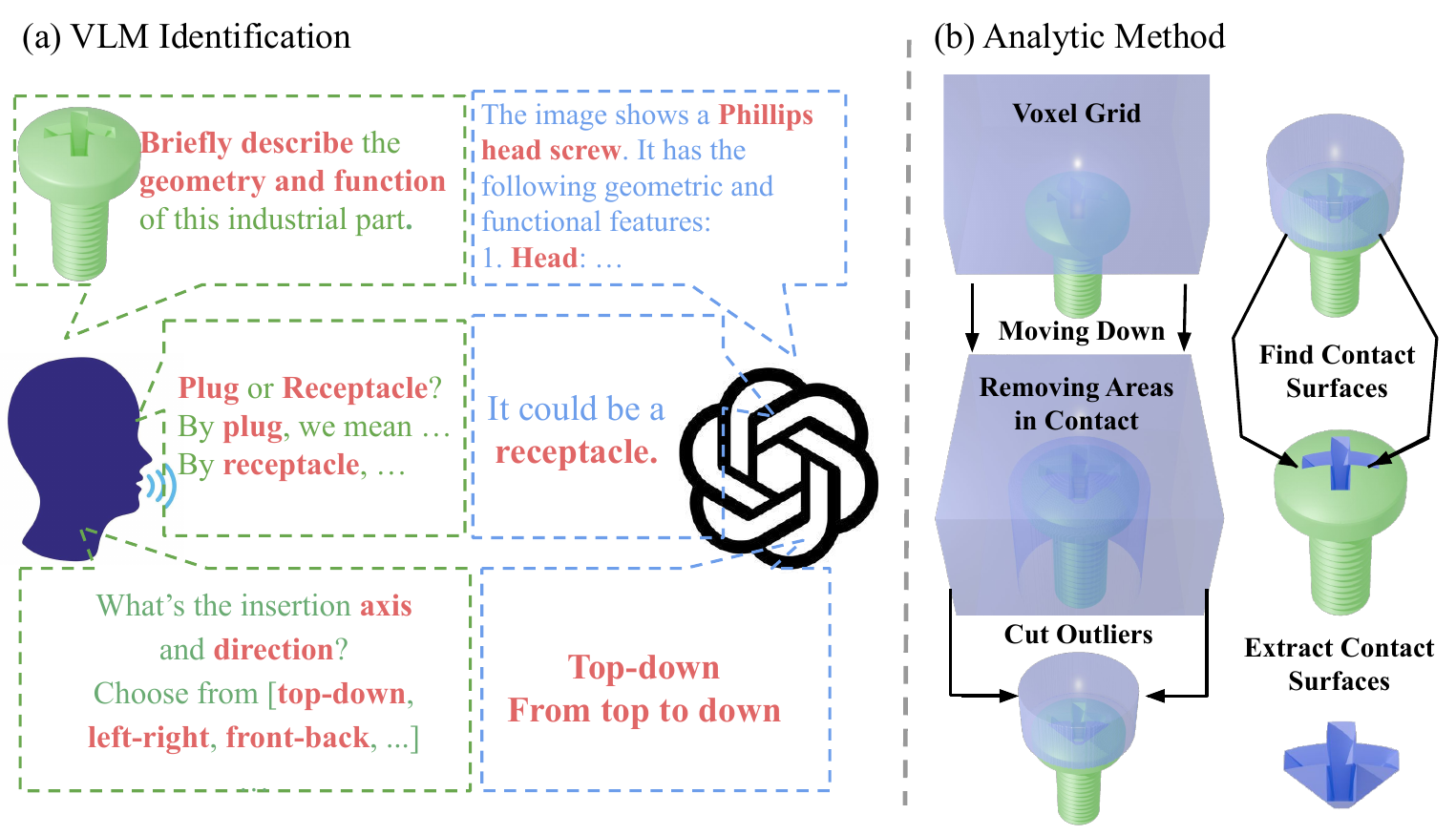}\qquad
\end{center}
\vspace{-6mm}
\caption{Contact-Surface Extraction: (a) A Vision-Language Model (VLM), such as GPT-4o, is employed to identify the object and predict potential insertion directions. (b) An analytical method is then applied to extract the contact surfaces based on the object's geometry and results from (a).}
\vspace{-2mm}
\label{fig:method}  
\end{figure}

Contact surfaces are extracted in two stages. First, we render an initially sampled or generated asset $A$ using Blender \cite{blender} and input it into the vision-language model GPT-4o to identify whether the asset is a plug or receptacle, as well as its assembly axis. To enhance asset recognition, we use chain-of-thought prompting to guide the VLM. Specifically, asset $A$ is normalized to fit within a $[-1, 1]^3$ bounding box and rendered from a top-front view at coordinates $(5, 5, 0)$ looking towards  $(0, 0, 0)$. The VLM is first prompted for a brief description of the asset in the image, followed by questions to determine if it is a plug or receptacle and the assembly direction when mated with a new asset.
For example, as shown in Fig.~\ref{fig:method} (a), the rendered asset is identified as a receptacle, with the assembly direction being top-down for a plug to mate into the screw head. 

There can be multiple plausible assembly axes. For example, a cylindrical plug is typically inserted along its height (top-down) but can also be inserted from the side, like a battery. The VLM samples one of these valid options. In each case, the second stage generates a corresponding receptacle aligned with the chosen assembly direction.

In the second stage, an analytical procedure similar to 3D flood-filling uses the assembly-direction information to identify which faces of the CAD model $A$ will be in contact during assembly. As shown in Fig.~\ref{fig:method} (b), the procedure begins by rotating the asset to align the assembly direction with the z-axis in a ``top-down'' orientation. A $512^3$ voxel grid is then initialized above asset $A$ and moved down until the top grid layer aligns with the top surface of $A$. As the grid descends, any cells that contact asset $A$ are removed. 
For receptacles, we also remove grid cells outside the convex hull of $A$.
Finally, we iterate through the faces in the B-rep representation of asset $A$ to identify which are fully in contact with the remaining portion of the voxel grid. This process determines the contact surfaces on asset $A$ that will be shared with the second mating asset that is to be generated. 
When generating a plug, we remove the top surfaces of the receptacle, retaining only the faces of the assembly orifices extending into the receptacle. This allows for greater variation in the generated asset's structure.

\subsection{Shape Completion}

Given the contact surfaces extracted from the previous stage, the objective is to generate a new asset that incorporates these contact surfaces while completing the external surface geometry of the generated part. To achieve this, we generate the new mesh in the B-rep format, which is a widely used standard in CAD models due to its smooth surface representation and robustness.
The B-rep format is typically defined as a graph where geometric primitives (faces and edges) serve as the graph nodes, and their topological relationships are captured by the graph edges, combining to form a solid model.

Leveraging prior work BrepGen \cite{xu2024brepgen} and Repaint \cite{lugmayr2022repaint}, we are able to preserve the shared contact surfaces while generating the complementary geometry of the new asset. Specifically, BrepGen is a pipeline designed to generate 3D CAD models in B-rep format, utilizing latent diffusion models to generate faces, edges, and vertices, followed by post-processing to achieve the final B-rep structure. As noted in BrepGen, it can also perform shape completion, a process further enhanced by repainting techniques \cite{lugmayr2022repaint}.

To encourage the correct positioning of the generated asset $B$ (above asset $A$ and without enveloping the contact surfaces), we first move the contact surfaces to the bottom of the unit box in the final step of the Contact-Surface Extraction process (as shown in Fig.~\ref{fig:method} (b)). This step aligns with the training framework of the diffusion model, which generates assets within a unit box. We then apply the diffusion-based shape completion process from BrepGen to create the final geometry of the generated asset.

\subsection{Clearance Specification}

In the previous two stages, we generated asset $B$, which pairs with asset $A$ to form the assembly. However, asset $B$ may sometimes contain extra geometric features from the generation process that overlap with asset $A$. In addition, the generated shape may lack clearance between the parts, resulting in unstable behavior during simulation as shown in Fig. 1C. Clearance, defined as the minimum distance $c$ between the faces of the two assets in their assembled state, is crucial for simulator stability and realism. To address these issues, we apply a post-processing method to filter out the invalid portions of the assets and add the necessary clearance to the valid portions, making them suitable for simulation.

Specifically, similar to the analytical method described in Section \ref{section:A}, we first convert the generated CAD model into an occupancy grid representation. Next, initializing in the assembled state, we move this occupancy grid along the assembly axis until the parts completely separate. During this process, we remove the grids that come into contact with or fall within the clearance distance $c$ of the other asset. Finally, we use the marching cubes algorithm to convert the occupancy grid back into a triangular mesh to produce the final result.
This approach enables the generation of asset pairs with user-defined clearances in the assembled state, ensuring they remain free of intersections during assembly or disassembly. The ability to parameterize clearance is valuable for mesh augmentation during policy learning.

\section{Experiments}

In this section, we present experiments to demonstrate the efficiency of our generation pipeline, the quality of the generated results, and the value to contact-rich policy learning. Specifically, we evaluate the following aspects: (a) \textbf{Generation Efficiency}: the time required to generate an asset pair; (b) \textbf{Diversity}: the variation in generated results compared to existing datasets; (c) \textbf{Policy Learning}: the ability to simulate the generated results and train policies; and (d) \textbf{Real-World Deployment}: verification that the generated results can be successfully 3D-printed and assembled in the real world, and that simulation-trained policies with such assets can be deployed.

\subsection{Generation Efficiency}

In this section, we evaluate the generation efficiency of our method. Specifically, we evaluate 
the recognition success rate for \textit{Contact Surface Extraction}, the success rate and time cost for \textit{Shape Completion}, and the time cost for \textit{Clearance Specification}.

\paragraph{Contact Surface Extraction} As the analytical method in this section relies on the recognition results from GPT-4o \cite{islam2024gpt}, we report its success rate. We selected 50 assets and manually labeled them for \textit{Semantic} (plug or receptacle), \textit{Axis} (assembly joint axis), and \textit{Orientation} (assembly direction) to form the test set. We then compared our prompting method with an ablated version without chain-of-thought reasoning, in which GPT-4o is prompted to output all the results at once based on the image input. Table~\ref{tab:1} shows that the chain-of-thought method significantly improves recognition accuracy.

\begin{table}[h!]
\centering
\caption{Recognition accuracy of vision language model.}
\vspace{-2mm}
\begin{tabular}{c||ccc}
\hline
 & Semantic & Axis & Orientation \\
\hline
Ours & \textbf{70} & \textbf{66} & \textbf{60} \\
Ours w/o chain-of-thought & 56 & 56 & 46\\
\hline
\end{tabular}
\label{tab:1}
\end{table}

\paragraph{Shape Completion}
The shape completion process is the most time-consuming part of the pipeline, as it involves both sampling and denoising. We also observe significant variation in sample efficiency (the ratio between the number of samples and valid generated asset pairs) across different assets. To illustrate this, we evaluate five assets and present the sample efficiency and time required to generate valid asset pairs in Table~\ref{tab:2}. The results indicate that the generation process becomes more challenging as the number of shared faces and the complexity of the geometry increase. 
The primary cause of invalid cases is the failure to produce a valid B-rep format, often due to mismatched vertices or edges.
Another common issue arises when the generated asset mirrors the original asset instead of forming a complementary counterpart.
Incorporating surface normals as input to the generative model could help distinguish between inward- and outward-facing surfaces, guiding the generation toward the intended complementary shape.

\begin{table}[h!]
\centering
\caption{
Efficiency of shape completion process.
}
\begin{tabular}{c||ccccc}
\hline
Asset ID & 10174 & 10144 & 11213 & 10117 & 10168 \\
\hline
Shared Faces & 2 & 4 & 12 & 13 & 9 \\
Sample Efficiency & 0.10 & 0.04 & 0.03 & 0.02 & 0.01\\
Time Cost (s) & 114 & 251 & 376 & 491 & 1070 \\
\hline
\end{tabular}
\label{tab:2}
\end{table}

\begin{figure}[h]
\begin{center}
\noindent\includegraphics[width=.95\linewidth]{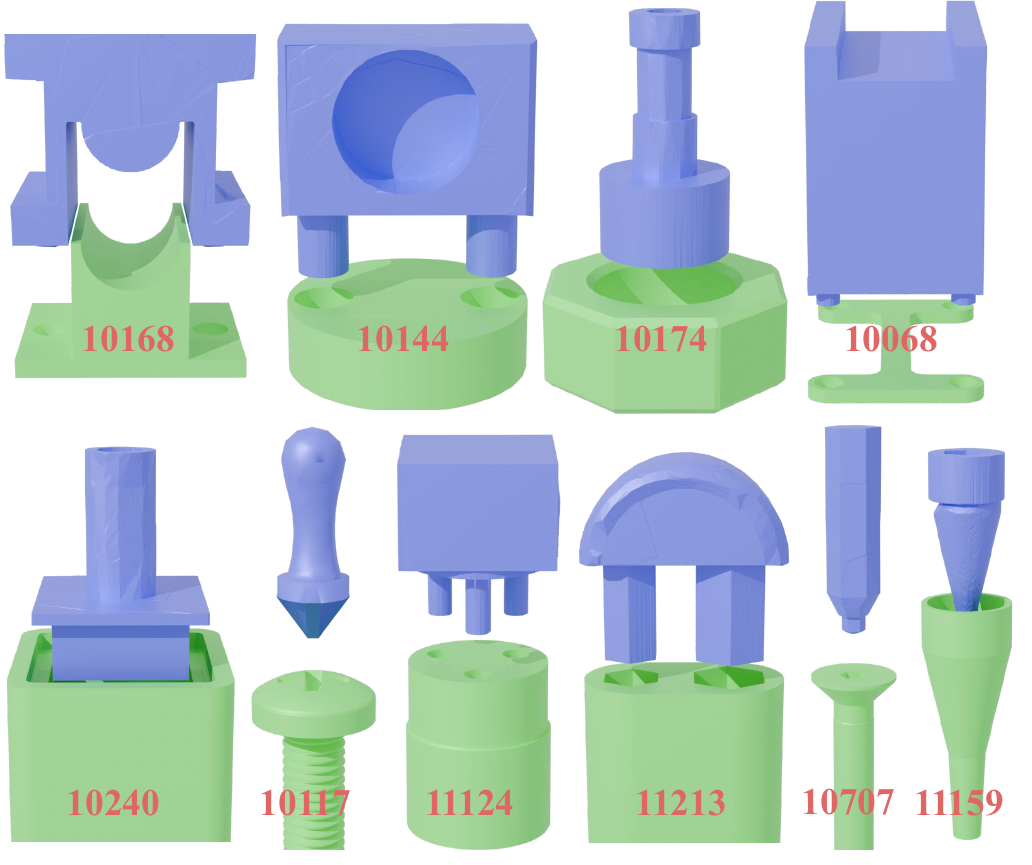}\qquad
\end{center}
\vspace{-4mm}
\caption{Samples of generated results. The green parts were sampled from the ABC dataset \cite{koch2019abc}, while the blue parts were generated. Asset pairs are labeled with unique identifiers (UIDs) for easy reference.}
\label{fig:results}  
\end{figure}

\begin{figure}[htb]
\begin{center}
\noindent\includegraphics[width=.95\linewidth, trim=0 0 0 60, clip]{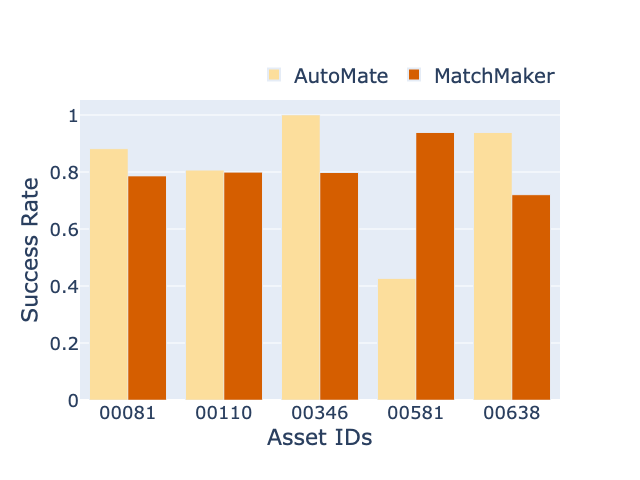}\qquad
\end{center}
\vspace{-8mm}
\caption{Policy-learning results of automatically post-processed assets (from {\sysName}) and manually post-processed assets (from AutoMate~\cite{tang2024automate}). For each, we select the best policy checkpoint and evaluate it across 3,000 trials.}
\vspace{-3mm}
\label{fig:1a}  
\end{figure}

\begin{figure}[h]
\begin{center}
\noindent\includegraphics[width=\linewidth, trim=0 10 30 97, clip]{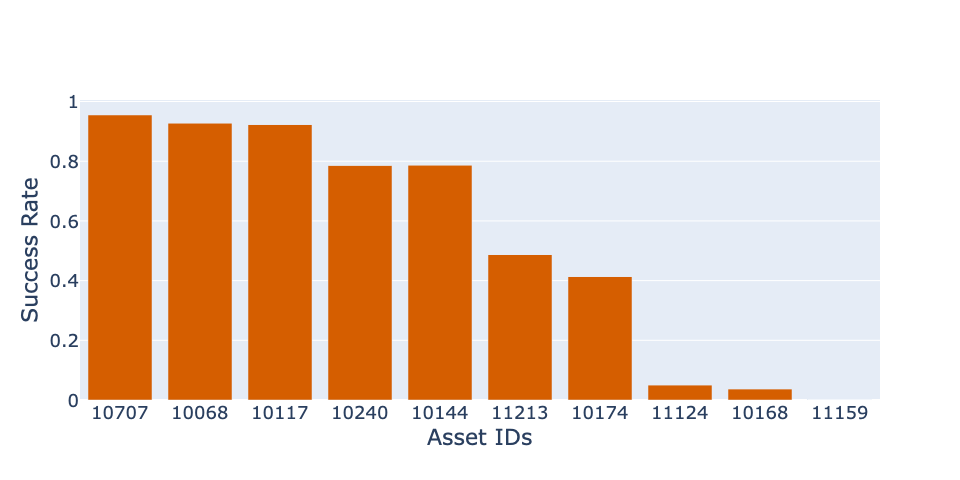}\qquad
\end{center}
\vspace{-8mm}
\caption{Policy-learning results of generated assets.}
\vspace{-4mm}
\label{fig:1b}  
\end{figure}

\paragraph{Clearance Specification}
The clearance specification process, which ensures collision-free results and maintains the desired clearance, %
typically takes around 120 seconds to post-process an asset pair at a $512^3$ resolution using 64 CPU cores. This represents a significant speedup compared to previous work \cite{tang2024automate}, which required approximately 40 hours of human effort to make 100 asset pairs collision-free and ensure proper clearance.

\subsection{Analysis of Generated Results}

Qualitatively, we present several examples of generated results in Fig.~\ref{fig:results}, showcasing the diversity and quality of our generated assets. Quantitatively, we evaluate the variance in grasp difficulty and shape complexity, and compare shape diversity across a fixed number of assets with previous works, including AutoMate \cite{tang2024automate} and the Assemble Them All dataset \cite{tian2022assemble}, as shown in Table~\ref{tab:3}.
To achieve the long-term goal of developing a generalist policy that works for many assemblies, it is essential to generate diverse assets with a wide range of complexity and grasp difficulty, covering both simple and complex objects. We measure these properties using methods from \cite{morrison2020egad}, computing grasp difficulty, shape complexity, and diversity scores for each object. These metrics are calculated for our generated assets and compared with those from two existing datasets.
Specifically, we evaluate the range of shape complexity scores (minimum to maximum), the range of grasp difficulty scores, and the overall shape diversity score.

For diversity, following \cite{morrison2020egad}, we define the similarity score for an asset as the average similarity with the 10 most similar assets in the dataset. The overall similarity score for a dataset is the average similarity score across all assets, and the diversity score is calculated as one minus the overall similarity score. To ensure a fair comparison, we greedily select 100 assets from each dataset to maximize the diversity score for comparison.

The results in Table~\ref{tab:3} show that our generated shapes exhibit a similar range of shape and grasp complexities compared to previous datasets, while achieving significantly higher diversity. We attribute this increased diversity, in part, to some characteristics of the Assemble Them All (ATA) dataset \cite{tian2022assemble}: 1) the plugs in the ATA dataset often have cylinder-like geometries, and 2) despite containing a total of 12,970 assembly assets, only 500 asset pairs involve single-axis insertion, which is the type of assembly we focus on.
In contrast, we generate assets with more diverse geometries and have already produced 1000 asset pairs in this study, with more continuously being generated.

\begin{table}[h!]
\centering
\caption{
Analysis of shape complexity, grasp difficulty, and diversity of different datasets using metrics from \cite{morrison2020egad}.%
}
\vspace{-2mm}
\begin{tabular}{c||cc|cc|c}
\hline
 & \multicolumn{2}{c|}{Shape Complexity} & \multicolumn{2}{c|}{Grasp Difficulty} & \multirow{2}{*}{Diversity $\uparrow$} \\
 & Min $\downarrow$ & Max $\uparrow$ & Min $\downarrow$ & Max $\uparrow$&  \\
\hline
Ours & 0.65 & \textbf{5.27} & \textbf{0.0} & \textbf{1.0} & \textbf{0.43} \\
AutoMate \cite{tang2024automate} & 0.59 & 5.04 & 0.12 & 0.87 & 0.19 \\
ATA \cite{tian2022assemble}& \textbf{0.0} & 4.98 & \textbf{0.0} & 0.93 & 0.24 \\
\hline
\end{tabular}
\label{tab:3}
\vspace{-3mm}
\end{table}

\subsection{Policy Learning}
We experimentally verify the effectiveness of {\sysName} for policy learning of contact-rich assembly tasks. This evaluation was conducted in two ways:

\begin{figure*}[htb]
\begin{center}
\noindent\includegraphics[width=.975\linewidth]{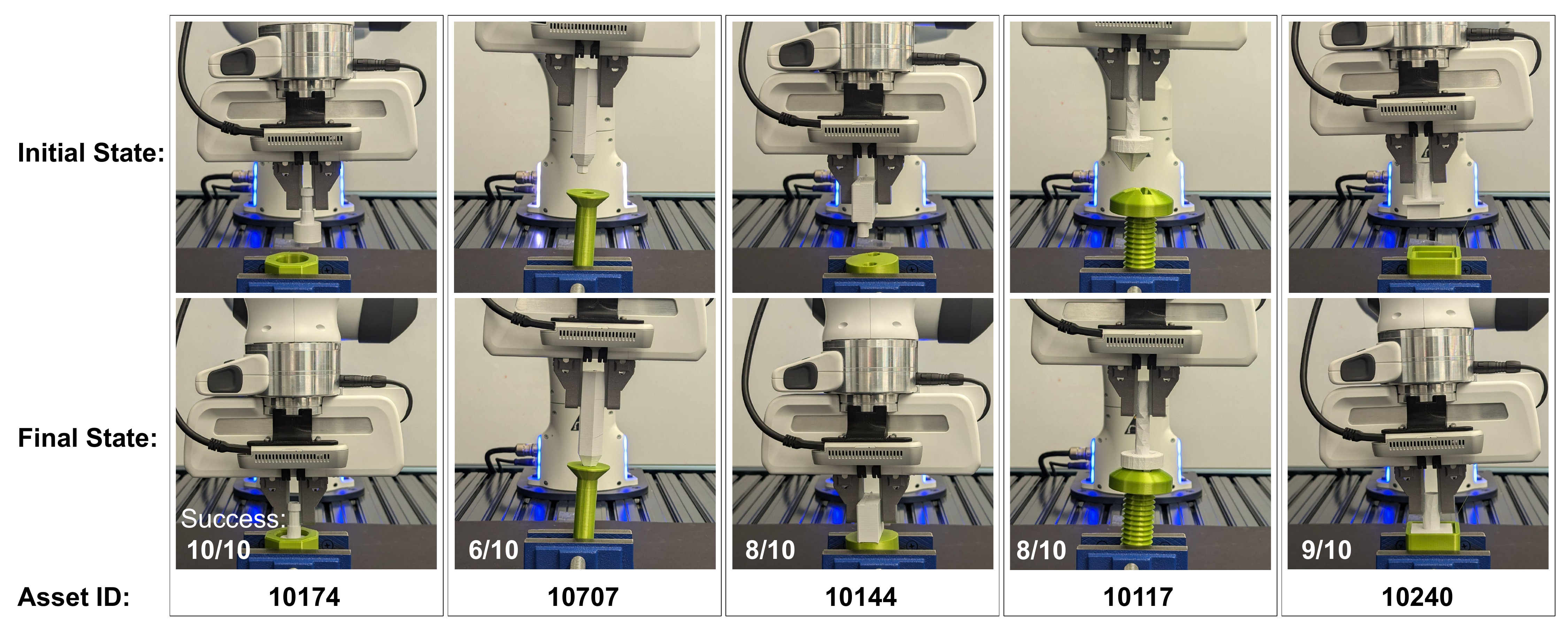}\qquad
\end{center}
\vspace{-4mm}
\caption{Key frames of the assembly process for each 3D-printed asset pairs, with the success rates over 10 trials. 
}
\vspace{-6mm}
\label{fig:real}  
\end{figure*}

\textbf{Asset Repair}: Here, we use the clearance-specification module of {\sysName} to repair assets from \cite{tian2022assemble} to automatically obtain the assets used in AutoMate \cite{tang2024automate}. The primary evaluation metric is the time required to perform the repair process. While the manual processing of 100 assets in \cite{tang2024automate} required over 40 hours of manual work, {\sysName} was able to repair the assets automatically in less than 1 minute per asset. We sampled 5 of the repaired assets and confirmed that we were able to learn a specialist policy for each asset with similar success rates as those reported in \cite{tang2024automate}  (Fig.~\ref{fig:1a}). This result indicates that the quality of the automatic repair process is comparable to manual efforts when evaluated based on the ultimate goal of policy-learning success.

\textbf{Asset Generation}: We also applied the policy-learning framework from \cite{tang2024automate} to a set of newly-generated assets and verified that assembly policies can also be obtained for these generated assets. The primary metric is whether the generative process can produce both easy and hard-to-solve assembly tasks. As shown in Fig.~\ref{fig:1b}, we can see that our generated assets can be simulated and learned by AutoMate \cite{tang2024automate}. 
Notably, some generated assets pose significant challenges for training specialist policies, creating new benchmarks for future assembly policy-learning algorithms. For instance, asset $10168$ in Fig. \ref{fig:results} has a geometry that requires more complex assembly behavior than AutoMate addresses, making it difficult to learn using the method. Similarly, asset $11159$ requires an assembly behavior that must account for the lack of exposed geometry for grasping in the assembled state, impeding the AutoMate method (which relies on first instantiating objects in the assembled state and generating demonstrations via grasping and disassembly).

\subsection{Real-world Demonstrations}

In addition to policy learning in simulation, we verify that assets generated by {\sysName} are realizable in the real world and that policies learned in simulation can transfer to assemble these parts. For this experiment, we 3D-print 5 generated asset pairs with high success rates in simulation and deploy the assembly policies in the real world (Fig.~\ref{fig:real}). 
At the start, a bench vise holds the receptacle and plug in the assembled state. For each trial, the robot grasps and lifts the plug, then randomly moves it to a new position before attempting to reinsert it. Compared to simulation, the receptacles in the real world are not entirely fixed and can move slightly during deployment.
We conducted 10 trials per asset, with the success rates shown in Fig.~\ref{fig:real}.
With an average success rate of 82\%, this experiment demonstrates that our generated assets can be used for both simulation and real-world policy benchmarking. The slightly higher success rates in real-world experiments likely result from the compliant fixturing of the socket, which can allow some degree of self-alignment during assembly.

\section{CONCLUSIONS}
In this work, we introduce {\sysName}, a pipeline for  generating diverse, simulation-compatible paired assets for robotic assembly policy learning. 
Using a diffusion-based shape completion backbone, our three-step pipeline (contact surface extraction, shape completion, and clearance specification) produces realistic assets with appropriate clearances for assembly tasks.
We validated {\sysName} in both simulated and real-world environments, demonstrating its effectiveness in developing robust assembly policies.
We plan to release the code from {\sysName}, allowing users to automatically repair interpenetrating BRep assemblies, specify desired clearances, generate new mating parts from input meshes, and instantiate RL environments for these assemblies within NVIDIA Isaac Lab.

\textbf{Limitations and Future Work:}
In addition to generating new datasets, our framework offers several data-augmentation possibilities:  
1) We can generate new plugs for existing receptacles and vice versa by extracting contact faces.
2) By using our clearance specification tool, we can create asset pairs with varying levels of clearance, which could enhance policy robustness to manufacturing tolerances and part variation.
3) For difficult asset pairs, we can increase clearance or generate similar, but diverse assets to help warm-start the learning process (e.g., as part of a curriculum-learning framework).

There are also a few limitations of this work: 
1) The complexity of the generated shapes is limited to 50 surfaces to stay within GPU memory capacity during training.
2) Generating a valid BRep asset can require multiple diffusion model samples; improved model architectures could increase the rate of valid BRep outputs.
3) Our approach mainly focuses on single-axis, two-part assembly tasks; future work will explore generating assemblies that require rotational or helical motions, or multiple assembly steps.

{
\bibliographystyle{IEEEtran}
\bibliography{IEEEabrv,references}
}

\end{document}